\documentclass[runningheads]{llncs}

\usepackage{eccv}

\usepackage{eccvabbrv}

\usepackage{graphicx}
\usepackage{booktabs}

\usepackage[accsupp]{axessibility}  

\usepackage{hyperref}

\usepackage{orcidlink}

\usepackage{multirow}
\usepackage[normalem]{ulem}
\useunder{\uline}{\ul}{}

\usepackage[stable]{footmisc}
\usepackage[LGR,T1]{fontenc}
\usepackage[utf8]{inputenc}
\usepackage{textalpha}

\begin{document}

\title{Flow4R: Unifying 4D Reconstruction and Tracking with Scene Flow}


\author{Shenhan Qian\inst{1,2}\orcidlink{0000-0003-0416-7548},
Ganlin Zhang\inst{1,2}\orcidlink{0009-0001-6089-8225},
Shangzhe Wu\inst{3,}$^\dagger$\orcidlink{0000-0003-1011-5963},
Daniel Cremers\inst{1,2,}$^\dagger$\orcidlink{0000-0002-3079-7984}
}

\begingroup
\renewcommand{\thefootnote}{}
\footnotetext{$^\dagger$Equal advising.}
\footnotetext{Project page: \url{https://shenhanqian.github.io/flow4r}}
\endgroup

\authorrunning{S.~Qian et al.}

\institute{Technical University of Munich \and
MCML \and 
University of Cambridge
}

\maketitle
\begin{abstract}
Reconstructing and tracking dynamic 3D scenes is a fundamental challenge in computer vision. 
Existing methods typically decouple geometry from motion: static multi-view reconstruction systems assume a rigid world, whereas dynamic tracking frameworks rely on explicit ego-motion estimation or separate object motion models. 
In this work, we propose Flow4R, a unified framework that treats relative scene flow as the central representation linking 3D structure, camera ego-motion, and dynamic object motion. 
Given a two-view input, Flow4R employs a shared Vision Transformer to predict a compact, pixel-aligned property set comprising 3D point positions, scene flow, pose weights, and confidence maps. 
This flow-centric formulation allows local geometry and bidirectional motion to be jointly inferred in a single feedforward pass, eliminating the need for explicit pose regression heads or complex bundle adjustment. 
By training jointly on static and dynamic datasets, Flow4R achieves state-of-the-art performance on 4D reconstruction and tracking benchmarks, demonstrating the power of the flow-centric formulation for spatiotemporal scene understanding.
\end{abstract}

\begin{figure}
    \centering
    \includegraphics[width=1\linewidth]{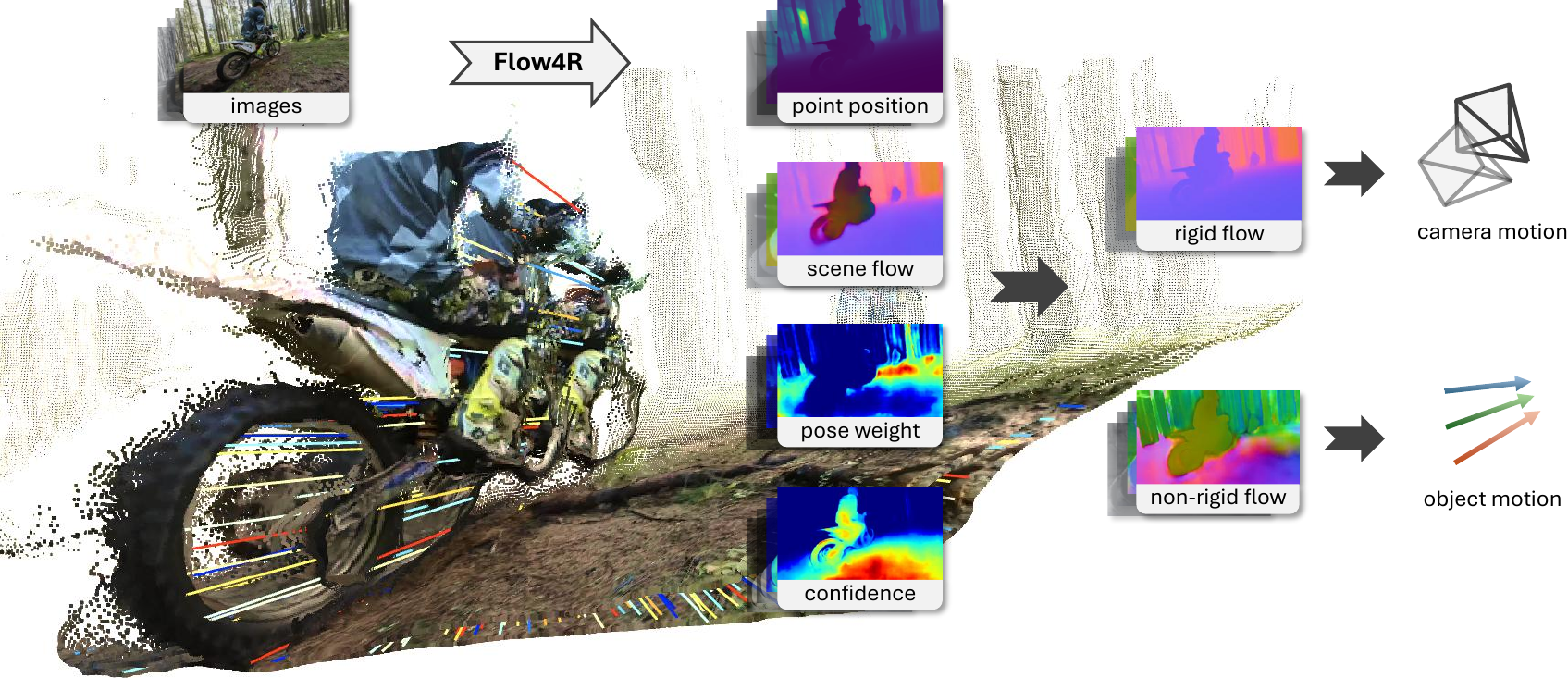}
    \caption{Given each image pair, Flow4R predicts for each image the point position $\mathbf{P}$, scene flow $\mathbf{F}$, pose weight $\mathbf{W}$, and confidence $\mathbf{C}$. Central to our framework, the scene flow $\mathbf{F}$ captures motion of points relative to the observer, thus is independent of the choice of coordinate system. Based on the pose weight $\mathbf{W}$, the scene flow $\mathbf{F}$ can be accurately decomposed into camera motion and object motion, enabling stable reconstruction and flexible tracking in both static and dynamic scenarios.}
    \label{fig:teaser}
\end{figure}
    
\section{Introduction}

\subsection{A Unified Representation for Dynamic Scenes}
Understanding dynamic 3D scenes requires reasoning jointly about \textit{geometry}, \textit{motion}, and \textit{time}. 
Yet existing methods typically treat these factors separately: multi-view reconstruction systems assume static worlds with fixed camera poses, while motion estimation and tracking frameworks rely on explicit pose regression or per-object motion models. 
This separation leads to brittle pipelines that struggle to generalize across diverse scenes and motion types. 
We argue that a single, unified representation---\textbf{scene flow}---can encapsulate both 3D structure and motion, bridging the gap between reconstruction and tracking. 

\begin{figure*}[t]
  \centering
   \includegraphics[width=1\linewidth]{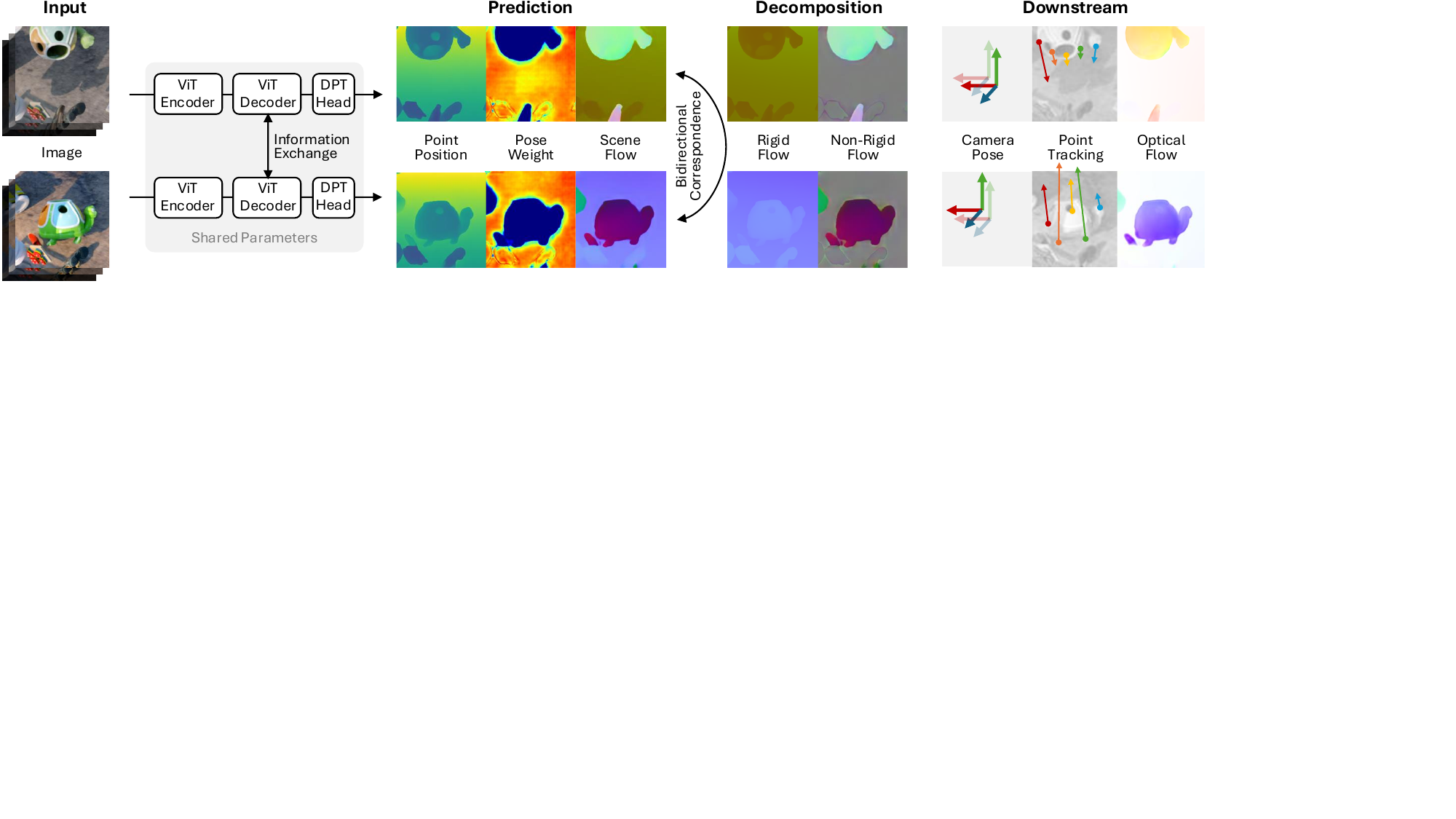}
   \caption{\textbf{Flow4R} takes two images as input at a time and predicts a pixel-aligned property set, including point position, scene flow, pose weight, and confidence, from which various downstream predictions can be deduced.}
   \label{fig:pipeline}
\end{figure*}

In this work, we introduce \textbf{Flow4R}---see Figures \ref{fig:teaser} and \ref{fig:pipeline}---a two-view transformer framework that formulates 4D perception entirely in terms of scene flow. 
Flow4R predicts per-pixel 3D positions, flow vectors, and camera pose weights, allowing it to infer both camera and object motion in a single forward pass without explicit pose heads or bundle adjustment. 
This flow-centric perspective reframes 4D reconstruction as continuous motion reasoning, unifying static and dynamic scene understanding under one compact formulation.

Recent advances in 3D reconstruction have seen the emergence of multi-view, self-attention-based transformers. 
DUSt3R~\cite{wang2023dust3r} reconstructs geometry by regressing pointmaps directly from image pairs using a transformer, while VGGT~\cite{wang2025vggt} generalizes this idea by handling multiple views jointly through global attention across all input images.
Subsequent work extends DUSt3R and VGGT to dynamic scenes by fine-tuning them on motion-rich datasets, with some methods incorporating correspondence supervision through optical flow or point tracking~\cite{st4rtrack2025, zhang2025pomato}. These approaches retain the assumption that all points are predicted in a shared reference frame---a natural choice for static scenes but potentially ambiguous in dynamic scenarios.

\subsection{Panta Rhei --- Everything Flows\footnote{Panta Rhei 
(\textalpha{πάντα ῥεῖ}), 
meaning ``everything flows'', is an aphorism which describes the doctrine of Heraclitus (520--460 BC).}}

Motion is relative: the observed motion in an image results from both the movements of objects and the ego-motion of the observer. 
When estimating camera motion from a dynamic sequence, we typically assume the scene is static and mask out dynamic regions. 
However, declaring a region as static is ultimately a choice of reference coordinate system. In many cases, the ground serves as a natural reference coordinate frame. However, this choice becomes ambiguous in dynamic scenes.
For instance, when approaching a moving escalator, our mind naturally treats the first step as static, but as we step off, the landing becomes the static reference frame. 
This illustrates that the selection of a reference frame is task-dependent and subjective. Therefore, a more general solution for 4D reconstruction and tracking would be to predict relative motion. Scene flow is a natural fit here since it is defined in the camera space, hence invariant to the choice of reference frame.

However, a key hurdle in training scene flow models is the extreme scarcity of ground-truth annotations for dynamic scenes~\cite{wedel2008efficient,wedel2011stereo,vogel2013piecewise,mayer2016large}. Moreover, most existing datasets are synthetic~\cite{mayer2016large,mehl2023spring}, which are accurate but limited in domain.
To address this, we propose a joint supervision strategy that leverages both static and dynamic datasets. 
For static scenes, which constitute the majority of available 3D data, we compute a \textit{rigid flow} using depth and relative poses to supervise the predicted flow. 
For dynamic scenes, we train the pose weight map $\mathbf{W}$ in a self-supervised, end-to-end manner by backpropagating through the weighted pose solver. 
This enables the network to automatically learn to segment static reference regions and down-weight dynamic, distant, or occluded pixels, directing the rigid motion supervision only to stable static structures. 
During inference, this learned weight map can be flexibly adjusted or overridden to switch reference coordinate frames based on the downstream task.

Furthermore, our two-view formulation scales naturally to sequential data. 
By pairing a sequence of frames with a single anchor frame, Flow4R performs bidirectional tracking and local geometry reconstruction. 
Unlike prior works, predicting local point maps for both views allows us to align the metric scale across arbitrary frame pairs using the shared anchor, enabling consistent 3D trajectory tracking without complex global bundle adjustment.

In summary, our contributions include:
\begin{itemize}
    \item A compact formulation under the two-view setting that supports various 4D tasks and generalizes to sequential data.
    \item An effective supervision strategy to learn local geometry and relative motion in a unified manner, leveraging both static and dynamic datasets.
    \item An efficient transformer-based model that achieves competitive performance for 4D reconstruction and tracking.
\end{itemize}
\section{Related Work}

\subsection{Structure-from-Motion (SfM)}
Classical Structure-from-Motion (SfM) methods~\cite{schonberger2016colmap, moulon2016openmvg,sweeney2015theia,pan2024glomap} typically rely on a modular three-stage pipeline comprising keypoint extraction and matching, incremental triangulation and registration, and bundle adjustment. These approaches utilize multi-view geometric constraints to explicitly optimize both 3D points and camera poses, yielding accurate and robust reconstructions. 
Later works~\cite{lindenberger2021pixel, wang2024vggsfm} make the pipeline fully differentiable, training the components end-to-end to improve robustness.
However, their reliance on iterative bundle adjustment optimizations limits their inference speed, particularly at scale, and they are typically constrained to sparse reconstructions.

DUSt3R~\cite{wang2023dust3r} pioneered feedforward 3D reconstruction using transformer-based architectures. It directly regresses pixel-aligned pointmaps for two input views in a shared coordinate frame, enabling dense reconstruction without explicit feature matching or optimization. 
Subsequent follow-up works extend this paradigm to more than two views. Spann3R~\cite{wang2025spann3r} and related methods~\cite{wang2025cut3r,cabon2025must3r} extend DUSt3R to sequential multi-view data, while VGGT~\cite{wang2025vggt} and similar approaches~\cite{elflein2025light3r,yang2025fast3r,tang2025mvdust3r,wang2025pi3,keetha2026mapanything} process multiple images jointly to estimate consistent geometry. 
Unlike many of these methods that rely on dedicated regression heads for camera pose estimation, Flow4R solves for camera poses implicitly using the predicted scene flow and pose weight maps. Crucially, the pose weight map not only filters out dynamic regions but also indicates which static structures are most reliable for ego-motion estimation.

\subsection{Dynamic Reconstruction and Point Tracking}
Reconstructing dynamic scenes is significantly more challenging than static 3D reconstruction, as classical multi-view geometric constraints are violated by non-rigid motion. Pioneering works like DynamicFusion~\cite{newcombe2015dynamicfusion} utilize a canonical model to represent dynamic objects and track non-rigid deformations via a warp field. However, they rely on high-quality RGB-D inputs and are highly sensitive to initialization.

To improve robustness, several approaches~\cite{kopf2021robust,zhang2022structure, zhang2025monst3r, li2025megasam, chen2025batrack} integrate learning-based predictions with classical optimization. For instance, MonST3R~\cite{zhang2025monst3r} fine-tunes DUSt3R~\cite{wang2023dust3r} to regress pointmaps for dynamic scenes, which are subsequently refined via post-optimization. Similarly, MegaSAM~\cite{li2025megasam} and BA-Track~\cite{chen2025batrack} incorporate predicted motion priors into dense bundle adjustment systems for stable dynamic reconstruction.

Other recent architectures~\cite{zhang2025pomato,sucar2025dynamic,st4rtrack2025,han2025d2ust3r} extend DUSt3R~\cite{wang2023dust3r} to directly regress dynamic 3D points or point trajectories. For example, given two views at different timestamps, ZeroMSF~\cite{liang2025zero}, Dynamic Point Maps~\cite{sucar2025dynamic}, and POMATO~\cite{zhang2025pomato} introduce separate regression heads for different timestamps, whereas St4RTrack~\cite{st4rtrack2025} and D$^2$USt3R~\cite{han2025d2ust3r} repurpose the pointmap prediction head for the second frame. In contrast to these head-heavy formulations, Flow4R employs a single shared decoder and head to predict a compact set of unified properties representing both the 3D geometry and bidirectional motion.

Concurrent works extend the multi-view VGGT~\cite{wang2025vggt} framework to sequential inputs~\cite{zhuo2026streamvggt}, dynamic scenes~\cite{zhou2025page, hu2025vggt4d}, and 3D point tracking~\cite{xiao2025spatialtrackerv2, karhade2025any4d, wang20254d, zhang2025efficiently, sucar2026v, badki2026l4p, liu2025trace}, achieving spatial-temporal consistency through feedforward inference. While Flow4R focuses on the two-view setting, it remains compatible with and could be extended to such multi-view attention mechanisms.

Another line of research leverages off-the-shelf predictions. For example, DELTA~\cite{ngo2025delta} lifts 2D pixel tracks into 3D using depth to maintain spatial consistency over long sequences. Similarly, TAPIP3D~\cite{zhang2025tapip3d} introduces a world-centric 3D feature cloud to enable persistent tracking under substantial camera ego-motion or occlusions. Along the same lines, C4D~\cite{wang2025c4d} combines feedforward 3D reconstruction networks with an optimized correspondence matching system and a learned mobility mask to perform joint 4D reconstruction and tracking.

\section{Method}

In this section, we first define a compact set of pixel-aligned properties and demonstrate how it unifies 4D reconstruction and tracking. We then describe how this formulation enables a simplified network design and allows flexible processing of sequential data. Please refer to the supplementary material for a summary table of notations.

\subsection{A Compact Property Set for 4D Tasks}
\label{sec:property_set}

\paragraph{Formulation.}
Given a pair of images $\left( \mathbf{I}, \mathbf{I}' \right)$ captured from different viewpoints and timestamps, we define a compact property set $\mathcal{S}(\mathbf{I}, \mathbf{I}') = \left\{\mathbf{P}, \mathbf{F}, \mathbf{W}, \mathbf{C}\right\}$ for pixels in image $\mathbf{I}$ with respect to $\mathbf{I}'$, where:
$\mathbf{P}\in\mathbb{R}^{H\times W \times 3}$ is a point position map placing pixels in the local Euclidean space;
$\mathbf{F}\in\mathbb{R}^{H\times W \times 3}$ is a scene flow map that maps each point from $\mathbf{I}$ to $\mathbf{I}'$ under camera and object motion;
$\mathbf{W}\in (0,1)^{H\times W}$ is a pose weight map with $\sum_{i=1}^{HW}{\mathbf{W}^i}=1$, indicating which pixels are reliable for camera pose estimation;
$\mathbf{C}\in (1,\infty)^{H\times W}$ is a confidence map. 
Symmetrically, we define the property set $\mathcal{S}(\mathbf{I}', \mathbf{I}) = \left\{ \mathbf{P}', \mathbf{F}', \mathbf{W}', \mathbf{C}' \right\}$ for image $\mathbf{I}'$ with respect to $\mathbf{I}$. This property set forms a compact parameter space for neural network prediction while enabling flexible configurations for various downstream tasks.

\paragraph{Scene Flow Decomposition.}
Induced by both camera and object movements, a scene flow vector $\mathbf{F}^i$ maps a point $\mathbf{P}^i$ to its corresponding location in the other frame's coordinate system and timestamp:
\begin{equation}
    \mathbf{P}^i_{vt} = \mathbf{P}^i + \mathbf{F}^i,
    \label{eq:p_vt}
\end{equation}
where $i$ is the pixel index, and $v, t$ indicate a switch of view and time from $\mathbf{I}$ to $\mathbf{I}'$.
Using the pose weight map $\mathbf{W}$, we solve for the relative camera pose $\mathrm{T}$ via a weighted least-squares formulation:
\begin{equation}
  \hat{\mathrm{T}} = \arg\min_{\mathrm{T} \in SE(3)} \sum^{HW}_{i=1} 
  \mathbf{W}^i \|  \mathbf{P}^i_{vt} - \mathrm{T} \mathbf{P}^i \|_2 .
  \label{eq:cam_pose}
\end{equation}
Thus, each point $\mathbf{P}^i$ projected into the other view's coordinate frame is written as:
\begin{equation}
    \mathbf{P}^i_{v} = \hat{\mathrm{T}} \mathbf{P}^i .
  \label{eq:p_v}
\end{equation}
Finally, we decompose the scene flow vector $\mathbf{F}^i$ into a rigid component (camera motion)
\begin{equation}
    \mathbf{F}^i_v = \mathbf{P}^i_{v} - \mathbf{P}^i ,
    \label{eq:f_v}
\end{equation}
and a non-rigid component (object motion)
\begin{align}
    \mathbf{F}^i_t =  \mathbf{F}^i - \mathbf{F}^i_{v}.
\end{align}

\paragraph{3D Point Tracking.}
One can track the trajectory of each point $\mathbf{P}^i$ in its own reference frame using:
\begin{equation}
    \mathbf{P}^i_t = \hat{\mathrm{T}}^{-1}\mathbf{P}^i_{vt} = \hat{\mathrm{T}}^{-1} \left( \mathbf{P}^i + \mathbf{F}^i \right) .
    \label{eq:p_t}
\end{equation}
\paragraph{Focal Length.}
Using the local point map $\mathbf{P}$, we solve for the focal length in image $\mathbf{I}$ via:
\begin{equation}
    \hat{f}=\arg\min_{f}\sum_{i}{
        \left(
            \left\| \hat{\mathbf{p}}^i - \pi(f,\mathbf{c},\mathbf{P}^i) \right\|^2
        \right),
    }
    \label{eq:focal_length}
\end{equation}
where $\hat{\mathbf{p}}^i$ and $\mathbf{c}$ are the image coordinates of pixel $i$ and the optical center, respectively, and $\pi(f,\mathbf{c},\mathbf{P}^i)$ is the perspective projection function. We assume identical focal lengths for both axes.

\paragraph{Optical Flow.}
Given the estimated or ground-truth focal length $f$, we project both $\mathbf{P}^i$ and $\mathbf{P}^i_{vt}$ into the image plane:
\begin{equation}
    \mathbf{p}^i = \pi(f,\mathbf{c},\mathbf{P}^i).
    \label{pts2d}
\end{equation}
The optical flow is subsequently computed as:
\begin{equation}
  \mathbf{f}^i = \mathbf{p}^i_{vt} - \mathbf{p}^i .
  \label{eq:optic_flow}
\end{equation}

\begin{figure}[t]
  \centering
   \includegraphics[width=1\linewidth]{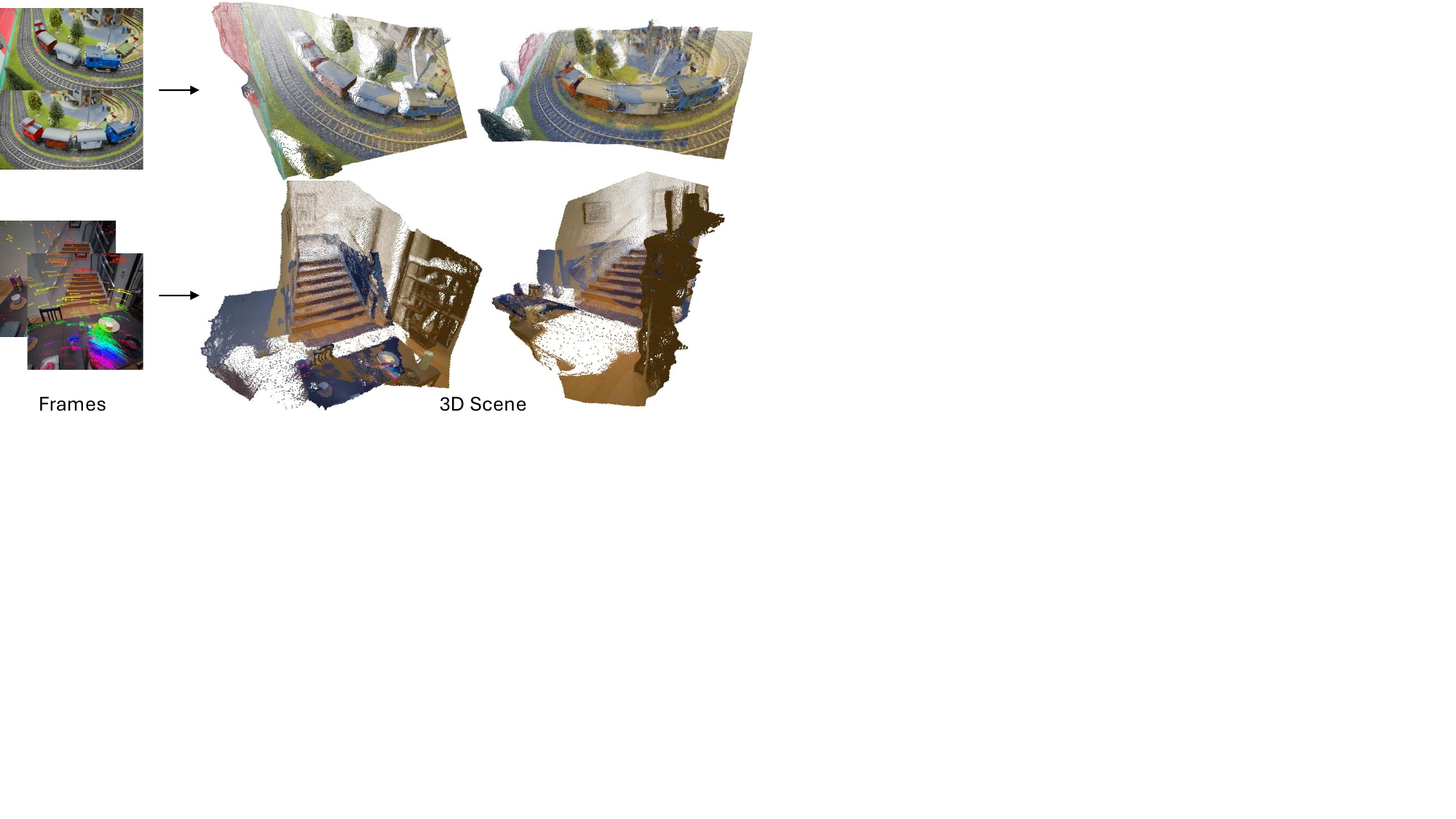}
   \caption{\textbf{3D Visualization of the Predictions.}  Examples are taken from the DAVIS~\cite{perazzi2016benchmark} and Aria Digital Twin~\cite{pan2023aria} datasets. Flow4R is capable of reconstructing 3D scenes and tracking the motion of both the camera and objects.}
   \label{fig:visual3d}
\end{figure}

\subsection{Flexible Predictions with a Compact Model}
\label{sec:model}
As shown above, the property set $\mathcal{S}$ is compact yet sufficient to recover local geometry and relative motion between two images, making it an ideal learning target for feedforward Vision Transformers.

\paragraph{Architecture Simplicity.}
Building on the success of two-view transformers like DUSt3R~\cite{wang2023dust3r} and MASt3R~\cite{leroy2024grounding}, which leverage cross-attention to match views and reconstruct geometry, we employ a similar architecture to learn relative scene flow. Thanks to our symmetric formulation, the twin forward paths share parameters across the encoders, decoders, and prediction heads. This symmetry also eliminates the need to manually construct symmetrized image pairs during training, like DUSt3R.

\paragraph{Inference Flexibility.}
During inference, the predicted property set $\mathcal{S}$ for each image encapsulates both local geometry and relative motion, supporting consistent 3D reconstruction, bidirectional tracking, and coordinate space transformations. In contrast, existing methods typically append dedicated prediction heads~\cite{sucar2025dynamic,zhang2025pomato} or repurpose static heads~\cite{han2025d2ust3r,st4rtrack2025}, which limits their flexibility when handling viewpoint or timestamp switches.

\begin{figure}[t]
  \centering
   \includegraphics[width=0.7\linewidth]{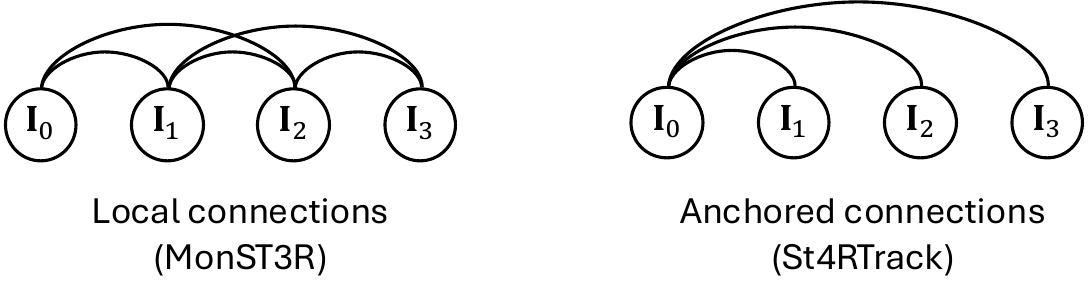}
   \caption{Sequence processing paradigms.}
   \label{fig:sequence_proc}
\end{figure}

\paragraph{Sequence Processing.}
The two-view formulation can be applied to video sequences by constructing image pairs, as shown in \cref{fig:sequence_proc}. MonST3R~\cite{zhang2025monst3r} constructs pairs within local temporal windows and uses post-optimization to arrange pairs in a pose graph. In contrast, St4RTrack~\cite{st4rtrack2025} builds anchored connections to conduct consistent tracking from the first image. To perform tracking and reconstruction without post-optimization, we adopt anchored connections following St4RTrack.

Because Flow4R predicts local point maps for both input views, it can align the metric scale of independent predictions using the shared anchor view. Specifically, given predictions for anchored pairs $\left[ \left( \mathbf{I}_0, \mathbf{I}_1 \right), \left( \mathbf{I}_0, \mathbf{I}_2 \right), \left( \mathbf{I}_0, \mathbf{I}_3 \right), \dots \right]$, we compute the average norm $s_n$ of the anchor point position map $\mathbf{P}_0^{(n)}$ predicted in each pair $n$. We then align the scale of the subsequent point maps and flows by scaling them with factors $\frac{s_1}{s_2}, \frac{s_1}{s_3}, \dots$. Crucially, this simple scale alignment is only possible because we predict the local point map of the anchor view in both forward paths, unlike previous methods that only output predictions for the target frame~\cite{st4rtrack2025,zhang2025pomato}.

\subsection{Model Training}
To train the network, we apply supervision on the predicted property set $\mathcal{S}$. Below we describe the loss formulations for $\mathcal{S}(\mathbf{I}, \mathbf{I}')$; the symmetric set $\mathcal{S}(\mathbf{I}', \mathbf{I})$ is supervised identically.

\paragraph{Normalization.}
Following DUSt3R~\cite{wang2023dust3r}, we normalize both the ground-truth and predicted point maps by their respective mean Euclidean norms to achieve scale-invariance. For the remainder of this section, we assume all 3D points ($\mathbf{P}$, $\mathbf{P}_v$, and $\mathbf{P}_{vt}$) are normalized.

\paragraph{Point Position Loss.}
Given the ground-truth point map $\bar{\mathbf{P}}$ obtained by back-projecting depth maps using camera intrinsics, we supervise the predicted point map $\mathbf{P}$ using a confidence-weighted regression loss:
\begin{equation}
    \mathcal{L}_{\mathbf{P}} = \frac{1}{\left| \mathbf{M}_{\mathbf{P}} \right|} \sum_{i}
        \mathbf{M}^i_{\mathbf{P}} \left(
            \mathbf{C}^i \| \mathbf{P}^i - \bar{\mathbf{P}}^i \|_2 - \alpha \log \mathbf{C}^i
        \right),
    \label{eq:l_P}
\end{equation}
where $\mathbf{M}_{\mathbf{P}}$ indicates pixels with valid depth ground truths.

\paragraph{3D Motion Loss.} 
In our formulation, scene flow is the key link across views and time. However, ground-truth scene flow annotations are scarce, particularly for dynamic scenes. To overcome this, we leverage diverse data sources providing pixel-wise correspondences on dynamic objects, including scene flow, optical flow, and 3D point tracking. While scene flow and optical flow are dense but restricted to adjacent frames, point tracking is sparser but provides long-term cross-frame correspondences. Fortunately, all these modalities can be converted into camera-space scene flow between the input frames. 

Empirically, we observe that directly predicting and supervising the point map $\mathbf{P}_{vt}$ yields superior performance compared to predicting the flow map $\mathbf{F}$ directly (\cref{tab:ablation}), despite their mathematical equivalence via $\mathbf{F} = \mathbf{P}_{vt} - \mathbf{P}$. Consequently, we define the direct network outputs as $\mathcal{P} = \{\mathbf{P}, \mathbf{P}_{vt}, \mathbf{W}, \mathbf{C}\}$ and adjust the loss formulations accordingly.

For ground-truth scene flow and 3D point tracking, we apply the 3D motion loss as:
\begin{equation}
    \mathcal{L}_{\mathbf{F}} = \frac{1}{\left| \mathbf{M}_{\mathbf{F}} \right|} \sum_{i}
        \mathbf{M}^i_{\mathbf{F}} \left(
            \mathbf{C}^i \| \mathbf{P}_{vt}^i - \bar{\mathbf{P}}_{vt}^i \|_2 - \alpha \log \mathbf{C}^i
        \right),
    \label{eq:l_F}
\end{equation}
where $\mathbf{M}^i_{\mathbf{F}}$ denotes the mask for pixels with ground-truth scene flows or tracked 3D points.
Since both the scene flow map and point position map are defined in Euclidean space, we share the confidence map between them for simplicity. Additionally, the confidence score dynamically increases as training proceeds, implicitly upweighting the scene flow loss along with the point position loss.

\paragraph{2D Motion Loss.}
When ground-truth optical flow or 2D point tracks are available, we apply the 2D motion loss to the projected coordinates $\mathbf{p}_{vt}$ (\cref{eq:optic_flow}):
\begin{equation}
    \mathcal{L}_{\mathbf{f}} = \frac{1}{\left| \mathbf{M}_{\mathbf{f}} \right|} \sum_{i}
        \mathbf{M}^i_{\mathbf{f}}
            \| \mathbf{p}_{vt}^i - \bar{\mathbf{p}}_{vt}^i \|_2 .
    \label{eq:l_f}
\end{equation}
We omit confidence weighting here since $\mathbf{p}_{vt}$ resides in 2D projective space rather than 3D Euclidean space.

\paragraph{Pose Weight Loss.}
Since scene flow captures the entirety of relative motion, camera ego-motion estimation requires identifying which regions conform to static scene geometry. This is modeled by the pose weight map $\mathbf{W}$. However, obtaining explicit ground-truth supervision for $\mathbf{W}$ is non-trivial. While foreground, instance, or rigidity masks are sometimes available, they do not perfectly represent optimal pose weights: foreground objects are useful for localization if static, and rigidity classifications can be ambiguous for slow-moving objects. Furthermore, $\mathbf{W}$ should ideally down-weight not only non-rigid motion but also distant, reflective, or occluded pixels that degrade pose solver stability.

To address this, we train the pose weight map in a self-supervised manner by backpropagating through the differentiable camera pose solver:
\begin{equation}
    \mathcal{L}_{\mathbf{W}} = \frac{1}{\left| \mathbf{M}_{\mathbf{P}} \right|} \sum_{i}
        \mathbf{M}^i_{\mathbf{P}} \left(
            \| \mathbf{P}^i_{v} - \bar{\mathbf{P}}^i_v \|_2
        \right),
    \label{eq:l_W}
\end{equation}
where $\mathbf{P}^i_v = \hat{\mathrm{T}} \mathbf{P}^i$, with $\hat{\mathrm{T}}$ being the camera pose estimated from $\mathbf{P}$, $\mathbf{P}_{vt}$, and $\mathbf{W}$ (\cref{eq:cam_pose,eq:p_v}), and $\bar{\mathbf{P}}^i_v = \bar{\mathrm{T}} \bar{\mathbf{P}}^i$, with $\bar{\mathrm{T}}$ representing the ground-truth relative camera pose. 
During backpropagation, we block gradients through $\mathbf{P}$ and $\mathbf{P}_{vt}$ since they are already supervised by other terms, making $\mathbf{W}$ the sole optimized parameter for this loss. Intuitively, this forces the network to adjust $\mathbf{W}$ to guide the solver towards the ground-truth ego-motion.

\paragraph{Rigid Motion Loss.}
Given the scarcity of dynamic scene flow annotations, we augment motion supervision by leveraging the rigid flow induced by the ground-truth camera pose. This is formulated as the rigid motion loss:
\begin{equation}
    \mathcal{L}_{\mathbf{F}_v} = \frac{1}{\left| \mathbf{M}_{\mathbf{P}} \right|} \sum_{i}
        \mathbf{M}^i_{\mathbf{P}} \left(
            w^i \mathbf{C}^i \| \mathbf{P}_{vt}^i - \bar{\mathbf{P}}^i_v \|_2 - \alpha \log \mathbf{C}^i
        \right),
    \label{eq:l_Fv}
\end{equation}
where:
\begin{equation}
w^i = 
\begin{cases} 
    sg(\mathbf{W}^i) \times HW & \text{if dynamic dataset},  \\
    1 & \text{if static dataset}.
\end{cases}
\end{equation}
For dynamic scenes, the predicted pose weight map $\mathbf{W}^i$ is used to down-weight the loss in non-rigid or dynamic regions. Since the weight map is normalized ($\sum_i^{HW} \mathbf{W}^i = 1$), we scale $\mathbf{W}^i$ by the image resolution $HW$ to maintain a loss magnitude consistent with other terms. We apply a stop-gradient operator $sg(\cdot)$ to prevent this loss from backpropagating into $\mathbf{W}$. For static datasets, $\mathbf{P}_{vt}$ is mathematically equivalent to the rigid projection $\mathbf{P}_v$, allowing direct regression to $\bar{\mathbf{P}}^i_v$ with $w^i = 1$.

\paragraph{Total Loss.} The multi-task loss is formulated as:
\begin{align}
    \mathcal{L}(\mathbf{P}, \mathbf{P}_{vt}, \mathbf{W}, \mathbf{C}) 
    &=
      \lambda_1 \mathcal{L}_{\mathbf{P}}
    + \lambda_2 \mathcal{L}_{\mathbf{F}} 
    + \lambda_3 \mathcal{L}_{\mathbf{f}} 
    + \lambda_4 \mathcal{L}_{\mathbf{W}}
    + \lambda_5 \mathcal{L}_{\mathbf{F}_v}
    \label{eq:l}
\end{align}
with $\lambda_1=1, \lambda_2=\lambda_4=\lambda_5=0.5, \lambda_3=0.3$, and $\alpha=0.2$ in \cref{eq:l_P,eq:l_Fv,eq:l_F}.

\section{Experiments}

\paragraph{Training Data.}
The training data is a combination of static and dynamic, real-world and synthetic datasets, including Habitat~\cite{habitat19iccv,szot2021habitat,puig2023habitat3}, BlendedMVS~\cite{yao2020blendedmvs}, MegaDepth~\cite{li2018megadepth}, ARKitScenes~\cite{baruch1arkitscenes}, CO3D~\cite{reizenstein2021common}, Static Scenes 3D~\cite{schroppel2022benchmark}, ScanNet++~\cite{yeshwanth2023scannet++}, Waymo~\cite{sun2020scalability}, TartanAir~\cite{wang2020tartanair}, UnReal4K~\cite{tosi2021smd}, WildRGBD~\cite{xia2024rgbd}, DL3DV~\cite{ling2024dl3dv}, MapFree~\cite{arnold2022map}, ScanNet~\cite{dai2017scannet}, HyperSim~\cite{roberts2021hypersim}, Virtual KITTI 2~\cite{cabon2020virtual}, Spring~\cite{mehl2023spring}, PointOdyssey~\cite{zheng2023pointodyssey}, Dynamic Replica~\cite{karaev2023dynamicstereo}, Kubric~\cite{greff2022kubric}, and OmniWorld-Game~\cite{zhou2025omniworld}. For some datasets, we use the data processing code or preprocessed data by DUSt3R~\cite{wang2023dust3r}, CUT3R~\cite{wang2025cut3r}, MonST3R~\cite{zhang2025monst3r}, and CoTracker~\cite{karaev2024cotracker}. Among the dynamic datasets, Virtual KITTI 2 provides ground-truth scene flow; Spring, Dynamic Replica, and OmniWorld-Game have ground-truth optical flow; PointOdyssey, Dynamic Replica, and Kubric contain 3D point tracking annotations.

\paragraph{Training Details.}
Unlike most of our baselines that are fine-tuned from DUSt3R~\cite{wang2023dust3r}, MASt3R~\cite{leroy2024grounding}, or MonST3R~\cite{zhang2025monst3r}, we initialize Flow4R from CroCo~\cite{croco, croco_v2} due to the formulation change and train it in two stages. In the first stage, we train the model with a linear head for 100 epochs on images at resolution 224. We sample 900K pairs for each epoch. In the second stage, we train the model with a DPT head for 100 epochs on images at resolution 512 with random aspect ratios. We sample 84K pairs for each epoch in the second stage. 
For video datasets, we first sample a random frame and then pair it with a neighbor within 50 frames of the same scene.
We use the Adam optimizer with linear learning rate warmup (10 epochs for the first stage, 20 epochs for the second stage) to reach the peak learning rate of 1e-4, then decay to 1e-6 until the end following a cosine curve. 
Gradients are clipped to a maximum norm of 10 with directions preserved.
We use a total batch size of 256 across eight NVIDIA A100/H100 GPUs for resolution 224 and a batch size of 64 for resolution 512. The entire training process takes around four days in total.

\begin{table}[t]
\caption{\textbf{World Coordinate 3D Point Tracking.} We report the performance on four datasets, Aria Digital Twin (ADT), Dynamic Replica (DR), Point Odyssey (PO), and Panoptic Studio (PS) using the Average Percentage of 3D Points within Delta (APD3D$\uparrow$) metric for all points and dynamic points after global median alignment. We also compare the model sizes in the last column. The best and second-best results are marked in bold and underlined.}
\label{tab:tracking}
\centering
\renewcommand{\arraystretch}{1.1}
\setlength{\tabcolsep}{7pt}
\begin{tabular}{@{}lcccccccc@{}}
\toprule
\multicolumn{1}{c}{} & \multicolumn{4}{c}{All Points} & \multicolumn{3}{c}{Dynamic Points} & \multirow{2}{*}{\begin{tabular}[c]{@{}c@{}}\# param. \\ (B)\end{tabular}} \\ \cmidrule(lr){2-5} \cmidrule(lr){6-8} 
\multicolumn{1}{c}{} & ADT & DR & PO & PS & ADT & DR & PO &  \\ \midrule
MonST3R & 74.4 & 58.1 & 33.5 & 51.3 & 67.9 & 51.9 & 39.4 & 0.7 \\
SpaTracker & 45.7 & 54.9 & 38.5 & 62.6 & 67.7 & 58.7 & 51.2 & \textbf{0.2} \\
POMATO & 57.2 & 68.4 & 49.7 & {\ul 64.9} & \textbf{78.1} & 62.7 & 58.1 & 0.7 \\
St4RTrack & {\ul 76.0} & {\ul 73.7} & {\ul 68.0} & \textbf{69.7} & {\ul 75.3} & {\ul 68.1} & {\ul 68.7} & 0.7 \\
\textbf{Flow4R} & \textbf{78.6} & \textbf{78.5} & \textbf{71.1} & 64.3 & 70.9 & \textbf{77.2} & \textbf{72.9} & {\ul 0.4} \\ \bottomrule
\end{tabular}
\end{table}

\subsection{Tracking and Reconstruction}
\paragraph{Benchmarks.}
Since we focus on reconstruction and tracking in a consistent coordinate system, we follow St4RTrack~\cite{st4rtrack2025} to evaluate in world coordinates using the WorldTrack benchmark. For 3D point tracking, WorldTrack comprises two real-world datasets, Aria Digital Twin~\cite{pan2023aria} and Panoptic Studio~\cite{joo2015panoptic}, as well as two synthetic datasets from the test sets of Point Odyssey~\cite{zheng2023pointodyssey} and Dynamic Replica~\cite{karaev2023dynamicstereo}, which provide 3D point trajectories. For dynamic 3D reconstruction, the synthetic Point Odyssey dataset and the real-world TUM-Dynamics~\cite{sturm2012benchmark} dataset are used for testing.

\paragraph{Metrics.}
Following St4RTrack~\cite{st4rtrack2025}, we adopt the Average Percentage of 3D Points within Delta (APD3D) metric for 3D point tracking evaluation. The predicted 3D point trajectories, after alignment to the ground truth, are compared frame by frame. We compute the prediction error and report the percentage of points whose error falls below a threshold $\delta_{3D}\in$ \{0.1m, 0.3m, 0.5m, 1.0m\}, averaged over the first 64 frames. For dynamic 3D reconstruction, we compare reconstructed point clouds against the ground truth using both APD3D and End-Point Error (EPE) metrics.

\begin{table}[t]
\caption{\textbf{World Coordinate 3D Reconstruction.} We report performance on Point Odyssey and TUM-Dynamics after
global median scaling. The best and second-best results are marked in bold and underlined.}
\label{tab:reconstruction}
\centering
\renewcommand{\arraystretch}{1.1}
\setlength{\tabcolsep}{9pt}
\begin{tabular}{@{}cccccc@{}}
\toprule
\textbf{} & \multicolumn{1}{c}{\textbf{}} & \multicolumn{2}{c}{Point Odyssey} & \multicolumn{2}{c}{TUM-Dynamics} \\ \cmidrule(l){3-4} \cmidrule(l){5-6} 
Category & \multicolumn{1}{c}{Method} & APD$\uparrow$ & EPE$\downarrow$ & APD$\uparrow$ & EPE$\downarrow$ \\ 
\midrule
\multirow{3}{*}{w/ Global Align.} & DUSt3R+GA & 43.90 & 0.609 & 70.49 & 0.315 \\
 & MASt3R+GA & 60.44 & 0.403 & 68.38 & 0.519 \\
 & MonST3R+GA & 72.31 & 0.263 & 63.87 & 0.343 \\ \midrule
\multirow{6}{*}{Feedforward} & DUSt3R & 45.79 & 0.639 & 72.26 & 0.289 \\
 & MASt3R & 56.90 & 0.464 & 66.22 & 0.551 \\
 & MonST3R & 68.25 & 0.304 & 61.38 & 0.365 \\
 & POMATO & 66.50 & 0.385 & 49.80 & 0.509 \\
 & St4RTrack & {\ul 78.73} & {\ul 0.205} & \textbf{83.42} & \textbf{0.185} \\
 & \textbf{Flow4R} & \textbf{81.00} & \textbf{0.182} & {\ul 79.87} & {\ul 0.202} \\ \bottomrule
\end{tabular}
\end{table}

\paragraph{Baselines.}
Flow4R is primarily compared with other feedforward tracking and reconstruction methods.
For 3D point tracking, we evaluate against a camera-coordinate 3D tracking method, SpatialTracker~\cite{xiao2024spatialtracker}, a dynamic 3D reconstruction method, MonST3R~\cite{zhang2025monst3r}, and two dynamic 3D tracking methods POMATO~\cite{zhang2025pomato} and St4RTrack~\cite{st4rtrack2025}. 
For POMATO, since its sequential model only provides tracking from other frames to the anchor frame (rather than the standard anchor-to-other-frame tracking used by other methods), we adopt its pairwise model for evaluation.
For 3D reconstruction evaluation, the static methods DUSt3R~\cite{wang2023dust3r} and MASt3R~\cite{leroy2024grounding} are also compared.

\paragraph{Results.}
In \cref{tab:tracking}, we conduct an evaluation on world-coordinate 3D point tracking, which is essentially $\mathbf{P}_t$ (\cref{eq:p_t}) for Flow4R. As the APD metric measures how accurately points are tracked in world space, a model needs to predict high-quality geometry, camera motion, and point movement at the same time. Flow4R shows higher performance on most datasets despite having fewer parameters. Note that all the baselines except SpaTracker follow the asymmetric head-bound formulation of DUSt3R, while Flow4R's competitiveness indicates the potential of our symmetric minimal formulation.

In \cref{tab:reconstruction}, we evaluate the world-space reconstruction quality via point maps in the reference view, which corresponds to $\mathbf{P}_v$ (\cref{eq:p_v}) in our formulation. Flow4R outperforms most baselines including those relying on post-optimization. This further validates the capacity and potential of our formulation with reconstruction and tracking unified by scene flows.

\begin{figure}[t]
  \centering
   \includegraphics[width=1\linewidth]{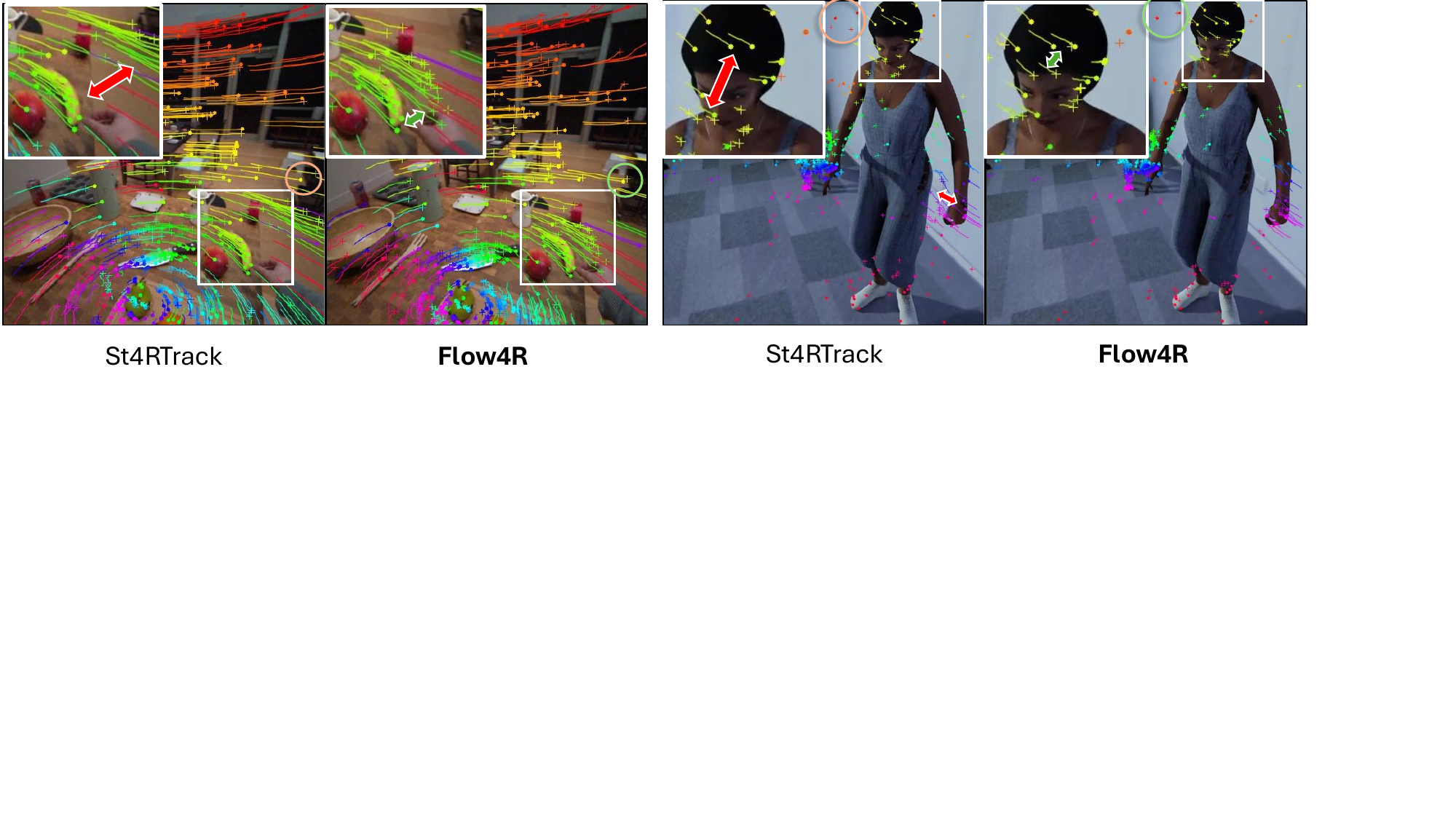}
   \caption{\textbf{Qualitative Results.} We visualize the 3D tracking trajectories by projecting them onto 2D. Ground-truth trajectories are marked with dots ($\bullet$), and predicted trajectories are denoted by a plus symbol ($+$). The results of Flow4R show less reprojection error on both the background and foreground.}
   \label{fig:comparison_tracking}
\end{figure}

As a qualitative comparison, we show sample results of the most competitive baselines in \cref{fig:comparison_tracking}. Flow4R predicts point positions precisely and tracks the motion of moving parts effectively.

\begin{table}[t]
\footnotesize
\centering
\renewcommand{\arraystretch}{1.1}
\setlength{\tabcolsep}{9pt}
\caption{\textbf{Ablation Study on Motion Representation.} We compare three variants of Flow4R using the Average Points under Distance (APD3D$\uparrow$) metric across all points. Our results demonstrate that directly predicting $\mathbf{P}_{vt}$ and supervising it with its corresponding ground truth yields the optimal performance across all datasets. For efficiency, these ablation results are reported from models trained for half the total number of epochs, which is sufficient to observe relative performance trends.}
\label{tab:ablation}
\begin{tabular}{@{}llcccccc@{}}
\toprule
\multirow{2}{*}{Pred.} & \multirow{2}{*}{Target} & \multicolumn{4}{c}{Tracking} & \multicolumn{2}{c}{Reconstruction} \\ \cmidrule(l){3-6} \cmidrule(l){7-8} 
 &  & ADT & DR & PO & PS & PO & TUM \\ \midrule
$\mathbf{F}$ & $\mathbf{\bar{F}}$ & {\ul 78.03} & 73.26 & 60.23 & 55.80 & {\ul 69.36} & 79.78 \\
$\mathbf{F}$ & $\mathbf{\bar{P}}_{vt}$ & 77.72 & {\ul 76.41} & {\ul 61.21} & {\ul 63.69} & 66.29 & {\ul 80.05} \\
$\mathbf{P}_{vt}$ & $\mathbf{\bar{P}}_{vt}$ & \textbf{78.50} & \textbf{78.48} & \textbf{67.93} & \textbf{67.17} & \textbf{77.20} & \textbf{80.34} \\ \bottomrule
\end{tabular}
\end{table}

\subsection{Ablation Study}
Given the compact property set $\mathcal{S}$ defined in \cref{sec:property_set}, a straightforward approach is to have the network directly predict the scene flow $\mathbf{F}$. However, as reconstruction and tracking accuracy are the primary objectives for real-world applications, we ablate three variants of Flow4R. Specifically, the network can predict either $\mathbf{F}$ or $\mathbf{P}_{vt}$---deriving the alternative via \cref{eq:p_vt}---while the regression target can be set to either $\mathbf{\bar{F}}$ or $\mathbf{\bar{P}}_{vt}$. Empirically, we find that directly predicting $\mathbf{P}_{vt}$ and regressing to $\mathbf{\bar{P}}_{vt}$ yields superior performance (\cref{tab:ablation}). This improvement likely stems from the fact that $\mathbf{P}_{vt}$ aligns more closely with the final evaluation metrics, which are based on absolute point positions. Conversely, $\mathbf{F}$ serves as an intermediate representation that requires additional transformations to compute final positions, potentially compounding errors. Crucially, predicting either $\mathbf{F}$ or $\mathbf{P}_{vt}$ remains fundamentally distinct from predicting $\mathbf{P}_{v}$ or $\mathbf{P}_{t}$, as both $\mathbf{F}$ and $\mathbf{P}_{vt}$ are invariant to the choice of reference objects within a dynamic scene.

\begin{figure}[t]
\centering
\includegraphics[width=1\linewidth]{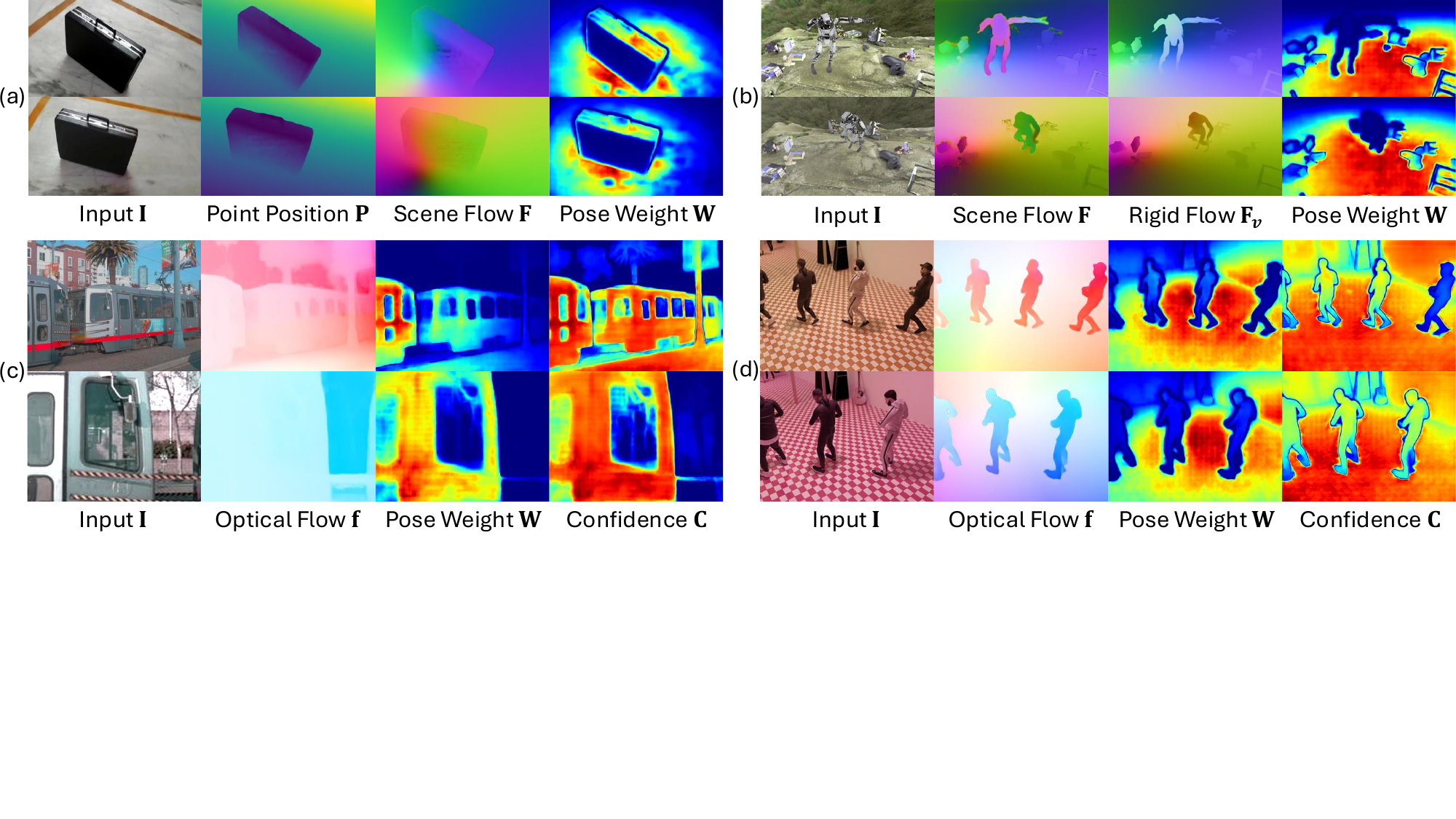}
\caption{
    \textbf{2D Visualization of Flow4R Predictions.} 
    Given each image pair, Flow4R predicts for each image the point position $\mathbf{P}$, scene flow $\mathbf{F}$, pose weight $\mathbf{W}$, and confidence $\mathbf{C}$. The point position map $\mathbf{P}$ captures scene geometry in the local space. The scene flow map $\mathbf{F}$ describes how each point moves from the current image to its pair, capturing both camera and object motions. The pose weight map $\mathbf{W}$ decides which pixels are reliable for camera pose estimation. The confidence map $\mathbf{C}$ indicates the uncertainty of the predictions. 
} 
\label{fig:visual2d}
\end{figure}

\subsection{Visualization}
To qualitatively analyze the internal behavior of Flow4R, we visualize its predictions across diverse scenarios. 
In \cref{fig:visual2d}~(a), which shows a static scene featuring a suitcase, the predicted scene flow exhibits a multi-directional color distribution, indicating rotation-dominated camera motion. 
Meanwhile, the pose weight map assigns low values to the dark, textureless regions of the suitcase, as these areas lack distinct features for localization. 
\cref{fig:visual2d}~(b) captures a dynamic environment with a jumping robot. 
Here, the scene flow aligns with the rigid flow map except in the region of the moving robot. 
Consequently, the pose weight map strongly suppresses the robot's pixels, preventing its independent motion from biasing the camera pose estimation. 
In \cref{fig:visual2d}~(c), which depicts a stationary train from both far and near perspectives, the optical flow maps---derived from predicted point positions and scene flow---are dominated by red and blue hues, signifying a primarily translational camera motion. 
Notably, the pose weight map isolates the overlapping regions between the two views, enhancing the interpretability of our camera pose estimation by explicitly identifying the pixels driving the calculation. 
Finally, \cref{fig:visual2d}~(d) records four individuals dancing. 
The confidence map reveals high certainty across the scene except for the last person in the line, who appears in only one frame. 
Simultaneously, the pose weight map excludes all moving figures, ensuring the camera pose is estimated solely from the static background.

In \cref{fig:visual3d}, we render the reconstructed and tracked points within a global coordinate system. Both dynamic elements (such as the train) and stationary structures (including the tree, ladder, wall, and table) exhibit high 3D consistency, demonstrating the effectiveness of our tracking and reconstruction pipeline.

\subsection{Runtime Efficiency} 
We further conduct a runtime efficiency analysis on an RTX PRO 6000 GPU. As shown in \cref{tab:runtime}, Flow4R achieves comparable throughput while saving more than 50\% VRAM compared to the most competitive baseline, St4RTrack.

\begin{table}[t]
\centering
\renewcommand{\arraystretch}{1.1}
\setlength{\tabcolsep}{9pt}
\caption{\textbf{Runtime efficiency metrics.} Flow4R achieves comparable throughput while saving more than 50\% VRAM compared to St4RTrack.}
\label{tab:runtime}
\begin{tabular}{ccc}
\toprule
                & Throughput (pairs/s)$\uparrow$  & VRAM (MB)$\downarrow$ \\
\midrule
St4RTrack       & 27.9                  & 6711 \\
\textbf{Flow4R} & 26.8                  & 3152 \\
\bottomrule
\end{tabular}
\end{table}
\section{Conclusion}
We presented Flow4R, a unified framework that establishes relative scene flow as the central representation linking 3D geometry, camera ego-motion, and dynamic object motion. By predicting a compact, coordinate-invariant property set—comprising 3D point positions, scene flow, pose weights, and confidence maps—Flow4R avoids separate pose regression heads or global bundle adjustment. Our joint training strategy successfully leverages both static and dynamic datasets, enabling self-supervised segmentation of static reference regions for robust pose estimation. We hope this flow-centric paradigm inspires further research towards holistic 4D perception, where reconstruction, tracking, and scene understanding emerge from a single coherent motion field.

\paragraph{Limitations.}
Despite its performance, Flow4R has several limitations that point to promising future directions. First, the performance of the model remains constrained by the limited availability of high-quality, real-world 3D scene flow data compared to static depth data. Second, our current formulation operates under a pairwise, two-view setup; extending the architecture to leverage multi-view attention mechanisms (e.g., as in VGGT~\cite{wang2025vggt}) could enhance temporal consistency and reconstructive fidelity. Finally, optimizing the framework for real-time online tracking under strict memory and computational budgets remains an important open challenge.

\section*{Acknowledgments}
This work was supported by the ERC Advanced Grant “SIMULACRON” (agreement \#884679), the GNI Project “AI4Twinning”, the DFG project CR 250/26-1 “4DYoutube", the Leibniz Supercomputing Centre (LRZ), and the UKRI AIRR programme. We would like to thank Weirong Chen, Dominik Muhle, and Linus Härenstam-Nielsen for valuable discussions throughout the project.

%
%
\bibliographystyle{splncs04}
\bibliography{main}

\begin{thebibliography}{10}
\providecommand{\url}[1]{\texttt{#1}}
\providecommand{\urlprefix}{URL }
\providecommand{\doi}[1]{https://doi.org/#1}

\bibitem{arnold2022map}
Arnold, E., Wynn, J., Vicente, S., Garcia-Hernando, G., Monszpart, A., Prisacariu, V., Turmukhambetov, D., Brachmann, E.: Map-free visual relocalization: Metric pose relative to a single image. In: European Conference on Computer Vision. pp. 690--708. Springer (2022)

\bibitem{badki2026l4p}
Badki, A., Su, H., Wen, B., Gallo, O.: {L4P}: {T}owards unified low-level {4D} vision perception. In: International Conference on 3D Vision (3DV) (2026)

\bibitem{baruch1arkitscenes}
Baruch, G., Chen, Z., Dehghan, A., Feigin, Y., Fu, P., Gebauer, T., Kurz, D., Dimry, T., Joffe, B., Schwartz, A., et~al.: Arkitscenes: A diverse real-world dataset for 3d indoor scene understanding using mobile rgb-d data. In: Thirty-fifth Conference on Neural Information Processing Systems Datasets and Benchmarks Track (Round 1) (2021)

\bibitem{cabon2020virtual}
Cabon, Y., Murray, N., Humenberger, M.: Virtual kitti 2. arXiv preprint arXiv:2001.10773  (2020)

\bibitem{cabon2025must3r}
Cabon, Y., Stoffl, L., Antsfeld, L., Csurka, G., Chidlovskii, B., Revaud, J., Leroy, V.: {MUSt3R}: Multi-view network for stereo {3D} reconstruction. In: CVPR. pp. 1050--1060 (2025)

\bibitem{chen2025batrack}
Chen, W., Zhang, G., Wimbauer, F., Wang, R., Araslanov, N., Vedaldi, A., Cremers, D.: Back on track: Bundle adjustment for dynamic scene reconstruction. In: Proceedings of the IEEE/CVF International Conference on Computer Vision (ICCV). pp. 4951--4960 (October 2025)

\bibitem{dai2017scannet}
Dai, A., Chang, A.X., Savva, M., Halber, M., Funkhouser, T., Nie{\ss}ner, M.: Scannet: Richly-annotated 3d reconstructions of indoor scenes. In: Proceedings of the IEEE conference on computer vision and pattern recognition. pp. 5828--5839 (2017)

\bibitem{elflein2025light3r}
Elflein, S., Zhou, Q., Leal-Taix{\'e}, L.: Light3r-sfm: Towards feed-forward structure-from-motion. In: Proceedings of the Computer Vision and Pattern Recognition Conference. pp. 16774--16784 (2025)

\bibitem{st4rtrack2025}
Feng*, H., Zhang*, J., Wang, Q., Ye, Y., Yu, P., Black, M.J., Darrell, T., Kanazawa, A.: {St4RTrack}: Simultaneous {4D} reconstruction and tracking in the world. In: Proceedings of the IEEE/CVF International Conference on Computer Vision (2025)

\bibitem{greff2022kubric}
Greff, K., Belletti, F., Beyer, L., Doersch, C., Du, Y., Duckworth, D., Fleet, D.J., Gnanapragasam, D., Golemo, F., Herrmann, C., et~al.: Kubric: A scalable dataset generator. In: Proceedings of the IEEE/CVF conference on computer vision and pattern recognition. pp. 3749--3761 (2022)

\bibitem{han2025d2ust3r}
Han, J., An, H., Jung, J., Narihira, T., Seo, J., Fukuda, K., Kim, C., Hong, S., Mitsufuji, Y., Kim, S.: Enhancing 3d reconstruction for dynamic scenes. In: The Thirty-ninth Annual Conference on Neural Information Processing Systems (2025)

\bibitem{hu2025vggt4d}
Hu, Y., Cheng, C., Yu, S., Guo, X., Wang, H.: Vggt4d: Mining motion cues in visual geometry transformers for 4d scene reconstruction. arXiv preprint arXiv:2511.19971  (2025)

\bibitem{joo2015panoptic}
Joo, H., Liu, H., Tan, L., Gui, L., Nabbe, B., Matthews, I., Kanade, T., Nobuhara, S., Sheikh, Y.: Panoptic studio: A massively multiview system for social motion capture. In: Proceedings of the IEEE international conference on computer vision. pp. 3334--3342 (2015)

\bibitem{karaev2023dynamicstereo}
Karaev, N., Rocco, I., Graham, B., Neverova, N., Vedaldi, A., Rupprecht, C.: Dynamicstereo: Consistent dynamic depth from stereo videos. In: Proceedings of the IEEE/CVF Conference on Computer Vision and Pattern Recognition. pp. 13229--13239 (2023)

\bibitem{karaev2024cotracker}
Karaev, N., Rocco, I., Graham, B., Neverova, N., Vedaldi, A., Rupprecht, C.: Cotracker: It is better to track together. In: European conference on computer vision. pp. 18--35. Springer (2024)

\bibitem{karhade2025any4d}
Karhade, J., Keetha, N., Zhang, Y., Gupta, T., Sharma, A., Scherer, S., Ramanan, D.: Any4d: Unified feed-forward metric 4d reconstruction. arXiv preprint arXiv:2512.10935  (2025)

\bibitem{keetha2026mapanything}
Keetha, N., M\"{u}ller, N., Sch\"{o}nberger, J., Porzi, L., Zhang, Y., Fischer, T., Knapitsch, A., Zauss, D., Weber, E., Antunes, N., Luiten, J., Lopez-Antequera, M., Bul\`{o}, S.R., Richardt, C., Ramanan, D., Scherer, S., Kontschieder, P.: {MapAnything}: Universal feed-forward metric {3D} reconstruction. In: International Conference on 3D Vision (3DV). IEEE (2026)

\bibitem{kopf2021robust}
Kopf, J., Rong, X., Huang, J.B.: Robust consistent video depth estimation. In: Proceedings of the IEEE/CVF Conference on Computer Vision and Pattern Recognition. pp. 1611--1621 (2021)

\bibitem{leroy2024grounding}
Leroy, V., Cabon, Y., Revaud, J.: Grounding image matching in {3D} with {MASt3R}. In: European Conference on Computer Vision. pp. 71--91. Springer (2024)

\bibitem{li2018megadepth}
Li, Z., Snavely, N.: Megadepth: Learning single-view depth prediction from internet photos. In: Proceedings of the IEEE conference on computer vision and pattern recognition. pp. 2041--2050 (2018)

\bibitem{li2025megasam}
Li, Z., Tucker, R., Cole, F., Wang, Q., Jin, L., Ye, V., Kanazawa, A., Holynski, A., Snavely, N.: Megasam: Accurate, fast and robust structure and motion from casual dynamic videos. In: Proceedings of the Computer Vision and Pattern Recognition Conference. pp. 10486--10496 (2025)

\bibitem{liang2025zero}
Liang, Y., Badki, A., Su, H., Tompkin, J., Gallo, O.: Zero-shot monocular scene flow estimation in the wild. In: Proceedings of the Computer Vision and Pattern Recognition Conference. pp. 21031--21044 (2025)

\bibitem{lindenberger2021pixel}
Lindenberger, P., Sarlin, P.E., Larsson, V., Pollefeys, M.: Pixel-perfect structure-from-motion with featuremetric refinement. In: Proceedings of the IEEE/CVF international conference on computer vision. pp. 5987--5997 (2021)

\bibitem{ling2024dl3dv}
Ling, L., Sheng, Y., Tu, Z., Zhao, W., Xin, C., Wan, K., Yu, L., Guo, Q., Yu, Z., Lu, Y., et~al.: Dl3dv-10k: A large-scale scene dataset for deep learning-based 3d vision. In: Proceedings of the IEEE/CVF Conference on Computer Vision and Pattern Recognition. pp. 22160--22169 (2024)

\bibitem{liu2025trace}
Liu, X., Xiao, Y., Chen, D.Y., Feng, J., Tai, Y.W., Tang, C.K., Kang, B.: Trace anything: Representing any video in 4d via trajectory fields. arXiv preprint arXiv:2510.13802  (2025)

\bibitem{mayer2016large}
Mayer, N., Ilg, E., Hausser, P., Fischer, P., Cremers, D., Dosovitskiy, A., Brox, T.: A large dataset to train convolutional networks for disparity, optical flow, and scene flow estimation. In: Proceedings of the IEEE conference on computer vision and pattern recognition. pp. 4040--4048 (2016)

\bibitem{mehl2023spring}
Mehl, L., Schmalfuss, J., Jahedi, A., Nalivayko, Y., Bruhn, A.: Spring: A high-resolution high-detail dataset and benchmark for scene flow, optical flow and stereo. In: Proceedings of the IEEE/CVF Conference on Computer Vision and Pattern Recognition. pp. 4981--4991 (2023)

\bibitem{moulon2016openmvg}
Moulon, P., Monasse, P., Perrot, R., Marlet, R.: Openmvg: Open multiple view geometry. In: International Workshop on Reproducible Research in Pattern Recognition. pp. 60--74. Springer (2016)

\bibitem{newcombe2015dynamicfusion}
Newcombe, R.A., Fox, D., Seitz, S.M.: Dynamicfusion: Reconstruction and tracking of non-rigid scenes in real-time. In: The IEEE Conference on Computer Vision and Pattern Recognition (CVPR) (June 2015)

\bibitem{ngo2025delta}
Ngo, T.D., Zhuang, P., Kalogerakis, E., Gan, C., Tulyakov, S., Lee, H.Y., Wang, C.: Delta: Dense efficient long-range 3d tracking for any video. In: The Thirteenth International Conference on Learning Representations (2025)

\bibitem{pan2024glomap}
Pan, L., Bar{\'a}th, D., Pollefeys, M., Sch{\"o}nberger, J.L.: Global structure-from-motion revisited. In: ECCV. pp. 58--77. Springer (2024)

\bibitem{pan2023aria}
Pan, X., Charron, N., Yang, Y., Peters, S., Whelan, T., Kong, C., Parkhi, O., Newcombe, R., Ren, Y.C.: Aria digital twin: A new benchmark dataset for egocentric 3d machine perception. In: Proceedings of the IEEE/CVF International Conference on Computer Vision. pp. 20133--20143 (2023)

\bibitem{perazzi2016benchmark}
Perazzi, F., Pont-Tuset, J., McWilliams, B., Van~Gool, L., Gross, M., Sorkine-Hornung, A.: A benchmark dataset and evaluation methodology for video object segmentation. In: Proceedings of the IEEE conference on computer vision and pattern recognition. pp. 724--732 (2016)

\bibitem{puig2023habitat3}
Puig, X., Undersander, E., Szot, A., Cote, M.D., Partsey, R., Yang, J., Desai, R., Clegg, A.W., Hlavac, M., Min, T., Gervet, T., Vondrus, V., Berges, V.P., Turner, J., Maksymets, O., Kira, Z., Kalakrishnan, M., Malik, J., Chaplot, D.S., Jain, U., Batra, D., Rai, A., Mottaghi, R.: Habitat 3.0: A co-habitat for humans, avatars and robots (2023)

\bibitem{reizenstein2021common}
Reizenstein, J., Shapovalov, R., Henzler, P., Sbordone, L., Labatut, P., Novotny, D.: Common objects in 3d: Large-scale learning and evaluation of real-life 3d category reconstruction. In: Proceedings of the IEEE/CVF international conference on computer vision. pp. 10901--10911 (2021)

\bibitem{roberts2021hypersim}
Roberts, M., Ramapuram, J., Ranjan, A., Kumar, A., Bautista, M.A., Paczan, N., Webb, R., Susskind, J.M.: Hypersim: A photorealistic synthetic dataset for holistic indoor scene understanding. In: Proceedings of the IEEE/CVF international conference on computer vision. pp. 10912--10922 (2021)

\bibitem{habitat19iccv}
Savva, M., Kadian, A., Maksymets, O., Zhao, Y., Wijmans, E., Jain, B., Straub, J., Liu, J., Koltun, V., Malik, J., Parikh, D., Batra, D.: Habitat: {A} {P}latform for {E}mbodied {AI} {R}esearch. In: Proceedings of the IEEE/CVF International Conference on Computer Vision (ICCV) (2019)

\bibitem{schonberger2016colmap}
Schonberger, J.L., Frahm, J.M.: Structure-from-motion revisited. In: Proceedings of the IEEE conference on computer vision and pattern recognition. pp. 4104--4113 (2016)

\bibitem{schroppel2022benchmark}
Schr{\"o}ppel, P., Bechtold, J., Amiranashvili, A., Brox, T.: A benchmark and a baseline for robust multi-view depth estimation. In: 2022 International Conference on 3D Vision (3DV). pp. 637--645. IEEE (2022)

\bibitem{sturm2012benchmark}
Sturm, J., Engelhard, N., Endres, F., Burgard, W., Cremers, D.: A benchmark for the evaluation of rgb-d slam systems. In: 2012 IEEE/RSJ international conference on intelligent robots and systems. pp. 573--580. IEEE (2012)

\bibitem{sucar2026v}
Sucar, E., Insafutdinov, E., Lai, Z., Vedaldi, A.: V-dpm: 4d video reconstruction with dynamic point maps. arXiv preprint arXiv:2601.09499  (2026)

\bibitem{sucar2025dynamic}
Sucar, E., Lai, Z., Insafutdinov, E., Vedaldi, A.: Dynamic point maps: A versatile representation for dynamic 3d reconstruction. In: Proceedings of the IEEE/CVF International Conference on Computer Vision (ICCV) (October 2025)

\bibitem{sun2020scalability}
Sun, P., Kretzschmar, H., Dotiwalla, X., Chouard, A., Patnaik, V., Tsui, P., Guo, J., Zhou, Y., Chai, Y., Caine, B., et~al.: Scalability in perception for autonomous driving: Waymo open dataset. In: Proceedings of the IEEE/CVF conference on computer vision and pattern recognition. pp. 2446--2454 (2020)

\bibitem{sweeney2015theia}
Sweeney, C., Hollerer, T., Turk, M.: Theia: A fast and scalable structure-from-motion library. In: ACM MM. pp. 693--696 (2015)

\bibitem{szot2021habitat}
Szot, A., Clegg, A., Undersander, E., Wijmans, E., Zhao, Y., Turner, J., Maestre, N., Mukadam, M., Chaplot, D., Maksymets, O., Gokaslan, A., Vondrus, V., Dharur, S., Meier, F., Galuba, W., Chang, A., Kira, Z., Koltun, V., Malik, J., Savva, M., Batra, D.: Habitat 2.0: Training home assistants to rearrange their habitat. In: Advances in Neural Information Processing Systems (NeurIPS) (2021)

\bibitem{tang2025mvdust3r}
Tang, Z., Fan, Y., Wang, D., Xu, H., Ranjan, R., Schwing, A., Yan, Z.: Mv-dust3r+: Single-stage scene reconstruction from sparse views in 2 seconds. In: Proceedings of the Computer Vision and Pattern Recognition Conference. pp. 5283--5293 (2025)

\bibitem{tosi2021smd}
Tosi, F., Liao, Y., Schmitt, C., Geiger, A.: Smd-nets: Stereo mixture density networks. In: Proceedings of the IEEE/CVF conference on computer vision and pattern recognition. pp. 8942--8952 (2021)

\bibitem{vogel2013piecewise}
Vogel, C., Schindler, K., Roth, S.: Piecewise rigid scene flow. In: Proceedings of the IEEE International Conference on Computer Vision. pp. 1377--1384 (2013)

\bibitem{wang20254d}
Wang, H., Zhou, H., Liu, H., Yan, L.: 4d-vggt: A general foundation model with spatiotemporal awareness for dynamic scene geometry estimation. arXiv preprint arXiv:2511.18416  (2025)

\bibitem{wang2025spann3r}
Wang, H., Agapito, L.: {3D} reconstruction with spatial memory. In: 3DV (2025)

\bibitem{wang2025vggt}
Wang, J., Chen, M., Karaev, N., Vedaldi, A., Rupprecht, C., Novotny, D.: Vggt: Visual geometry grounded transformer. In: Proceedings of the Computer Vision and Pattern Recognition Conference. pp. 5294--5306 (2025)

\bibitem{wang2024vggsfm}
Wang, J., Karaev, N., Rupprecht, C., Novotny, D.: Vggsfm: Visual geometry grounded deep structure from motion. In: Proceedings of the IEEE/CVF conference on computer vision and pattern recognition. pp. 21686--21697 (2024)

\bibitem{wang2025cut3r}
Wang, Q., Zhang, Y., Holynski, A., Efros, A.A., Kanazawa, A.: Continuous {3D} perception model with persistent state. In: CVPR. pp. 10510--10522 (2025)

\bibitem{wang2025c4d}
Wang, S., Jiang, Z., Yang, X., Wang, X.: C4d: 4d made from 3d through dual correspondences. In: Proceedings of the IEEE/CVF International Conference on Computer Vision. pp. 7570--7580 (2025)

\bibitem{wang2023dust3r}
Wang, S., Leroy, V., Cabon, Y., Chidlovskii, B., Revaud, J.: {DUSt3R}: Geometric {3D} vision made easy. 2024 ieee. In: CVF Conference on Computer Vision and Pattern Recognition (CVPR). pp. 20697--20709 (2023)

\bibitem{wang2020tartanair}
Wang, W., Zhu, D., Wang, X., Hu, Y., Qiu, Y., Wang, C., Hu, Y., Kapoor, A., Scherer, S.: Tartanair: A dataset to push the limits of visual slam. In: 2020 IEEE/RSJ International Conference on Intelligent Robots and Systems (IROS). pp. 4909--4916. IEEE (2020)

\bibitem{wang2025pi3}
Wang, Y., Zhou, J., Zhu, H., Chang, W., Zhou, Y., Li, Z., Chen, J., Pang, J., Shen, C., He, T.: $\pi^3$: Scalable permutation-equivariant visual geometry learning (2025), \url{https://arxiv.org/abs/2507.13347}

\bibitem{wedel2011stereo}
Wedel, A., Cremers, D.: Stereo scene flow for 3D motion analysis. Springer Science \& Business Media (2011)

\bibitem{wedel2008efficient}
Wedel, A., Rabe, C., Vaudrey, T., Brox, T., Franke, U., Cremers, D.: Efficient dense scene flow from sparse or dense stereo data. In: European conference on computer vision. pp. 739--751. Springer (2008)

\bibitem{croco_v2}
Weinzaepfel, P., Lucas, T., Leroy, V., Cabon, Y., Arora, V., Br{\'e}gier, R., Csurka, G., Antsfeld, L., Chidlovskii, B., Revaud, J.: {CroCo v2}: Improved cross-view completion pre-training for stereo matching and optical flow. In: ICCV (2023)

\bibitem{croco}
{Weinzaepfel, Philippe and Leroy, Vincent and Lucas, Thomas and Br\'egier, Romain and Cabon, Yohann and Arora, Vaibhav and Antsfeld, Leonid and Chidlovskii, Boris and Csurka, Gabriela and Revaud J\'er\^ome}: {CroCo}: Self-supervised pre-training for {3D} vision tasks by cross-view completion. In: {NeurIPS} (2022)

\bibitem{xia2024rgbd}
Xia, H., Fu, Y., Liu, S., Wang, X.: Rgbd objects in the wild: Scaling real-world 3d object learning from rgb-d videos. In: Proceedings of the IEEE/CVF Conference on Computer Vision and Pattern Recognition. pp. 22378--22389 (2024)

\bibitem{xiao2025spatialtrackerv2}
Xiao, Y., Wang, J., Xue, N., Karaev, N., Makarov, Y., Kang, B., Zhu, X., Bao, H., Shen, Y., Zhou, X.: Spatialtrackerv2: 3d point tracking made easy. In: Proceedings of the IEEE/CVF International Conference on Computer Vision (2025), \url{https://arxiv.org/abs/2507.12462}

\bibitem{xiao2024spatialtracker}
Xiao, Y., Wang, Q., Zhang, S., Xue, N., Peng, S., Shen, Y., Zhou, X.: Spatialtracker: Tracking any 2d pixels in 3d space. In: Proceedings of the IEEE/CVF Conference on Computer Vision and Pattern Recognition. pp. 20406--20417 (2024)

\bibitem{yang2025fast3r}
Yang, J., Sax, A., Liang, K.J., Henaff, M., Tang, H., Cao, A., Chai, J., Meier, F., Feiszli, M.: Fast3r: Towards 3d reconstruction of 1000+ images in one forward pass. In: Proceedings of the IEEE/CVF Conference on Computer Vision and Pattern Recognition (CVPR) (June 2025)

\bibitem{yao2020blendedmvs}
Yao, Y., Luo, Z., Li, S., Zhang, J., Ren, Y., Zhou, L., Fang, T., Quan, L.: Blendedmvs: A large-scale dataset for generalized multi-view stereo networks. In: Proceedings of the IEEE/CVF conference on computer vision and pattern recognition. pp. 1790--1799 (2020)

\bibitem{yeshwanth2023scannet++}
Yeshwanth, C., Liu, Y.C., Nie{\ss}ner, M., Dai, A.: Scannet++: A high-fidelity dataset of 3d indoor scenes. In: Proceedings of the IEEE/CVF International Conference on Computer Vision. pp. 12--22 (2023)

\bibitem{zhang2025tapip3d}
Zhang, B., Ke, L., Harley, A.W., Fragkiadaki, K.: Tapip3d: Tracking any point in persistent 3d geometry. In: The Thirty-ninth Annual Conference on Neural Information Processing Systems (2025)

\bibitem{zhang2025efficiently}
Zhang, C., Moing, G.L., Koppula, S., Rocco, I., Momeni, L., Xie, J., Sun, S., Sukthankar, R., Barral, J.K., Hadsell, R., et~al.: Efficiently reconstructing dynamic scenes one d4rt at a time. arXiv preprint arXiv:2512.08924  (2025)

\bibitem{zhang2025monst3r}
Zhang, J., Herrmann, C., Hur, J., Jampani, V., Darrell, T., Cole, F., Sun, D., Yang, M.H.: {MonST3R}: A simple approach for estimating geometry in the presence of motion. In: ICLR (2025)

\bibitem{zhang2025pomato}
Zhang, S., Ge, Y., Tian, J., Xu, G., Chen, H., Lv, C., Shen, C.: Pomato: Marrying pointmap matching with temporal motion for dynamic 3d reconstruction. In: Proceedings of the IEEE/CVF International Conference on Computer Vision (2025)

\bibitem{zhang2022structure}
Zhang, Z., Cole, F., Li, Z., Rubinstein, M., Snavely, N., Freeman, W.T.: Structure and motion from casual videos. In: European Conference on Computer Vision. pp. 20--37. Springer (2022)

\bibitem{zheng2023pointodyssey}
Zheng, Y., Harley, A.W., Shen, B., Wetzstein, G., Guibas, L.J.: Pointodyssey: A large-scale synthetic dataset for long-term point tracking. In: Proceedings of the IEEE/CVF International Conference on Computer Vision. pp. 19855--19865 (2023)

\bibitem{zhou2025page}
Zhou, K., Wang, Y., Chen, G., Beaudouin, G., Zhan, F., Liang, P.P., Wang, M.: Page-4d: Vggt-4d perception via disentangled pose and geometry estimation. In: The Fourteenth International Conference on Learning Representations

\bibitem{zhou2025omniworld}
Zhou, Y., Wang, Y., Zhou, J., Chang, W., Guo, H., Li, Z., Ma, K., Li, X., Wang, Y., Zhu, H., Liu, M., Liu, D., Yang, J., Fu, Z., Chen, J., Shen, C., Pang, J., Zhang, K., He, T.: Omniworld: A multi-domain and multi-modal dataset for 4d world modeling (2025), \url{https://arxiv.org/abs/2509.12201}

\bibitem{zhuo2026streamvggt}
Zhuo, D., Zheng, W., Guo, J., Wu, Y., Zhou, J., Lu, J.: Streaming 4d visual geometry transformer. In: The Thirteenth International Conference on Learning Representations (2026)

\end{thebibliography}


\section{Appendix}

This appendix starts with a compact reference of the symbols used throughout the paper and supplementary material. Table~\ref{tab:notations} summarizes the notation groups for inputs, outputs, and derived quantities to make subsequent descriptions easier to follow.

\begin{table}[h]
    \caption{Summary of notations.}
    \label{tab:notations}
    \centering
    \begin{tabular}{ll}
    \toprule
        \textbf{Symbol} & \textbf{Meaning} \\
        \midrule
        \multicolumn{2}{c}{Input} \\ \midrule
        $\mathbf{I} \in \mathbb{R}^{H\times W \times 3}$ & input image \\
        $\mathbf{I}' \in \mathbb{R}^{H\times W \times 3}$ & the other input image \\
        \midrule 
        \multicolumn{2}{c}{Output} \\ \midrule
        $\mathbf{P} \in \mathbb{R}^{H\times W \times 3}$ & point position map for $\mathbf{I}$ \\
        $\mathbf{F} \in \mathbb{R}^{H\times W \times 3}$ & scene flow map for $\mathbf{I}$ \\
        $\mathbf{W} \in \mathbb{R}^{H\times W}$ & pose weight map for $\mathbf{I}$ \\
        $\mathbf{C} \in \mathbb{R}^{H\times W}$ & confidence map for $\mathbf{I}$ \\
        \midrule 
        \multicolumn{2}{c}{Derived} \\ \midrule
        $\hat{\mathrm{T}} \in \mathbb{R}^{3\times 4}$ & solved rigid transformation of $\mathbf{I}$ relative to $\mathbf{I}'$\\
        $i \in \mathbb{N}$ & pixel index \\
        $\mathbf{P}^i \in \mathbb{R}^3$ & point position for pixel $i$ in $\mathbf{I}$ \\
        $\mathbf{P}^i_v \in \mathbb{R}^3$ & $\mathbf{P}^i$ in the view of $\mathbf{I}'$ \\
        $\mathbf{P}^i_t \in \mathbb{R}^3$ & $\mathbf{P}^i$ at the timestamp of $\mathbf{I}'$ \\
        $\mathbf{P}^i_{vt} \in \mathbb{R}^3$ & $\mathbf{P}^i$ in the view and at the timestamp of $\mathbf{I}'$ \\
        $\mathbf{F}^i \in \mathbb{R}^3$ & scene flow for pixel $i$ in $\mathbf{I}$ towards $\mathbf{I}'$ \\
        $\mathbf{F}^i_v \in \mathbb{R}^3$ & rigid component (camera motion) of $\mathbf{F}^i$ \\
        $\mathbf{F}^i_t \in \mathbb{R}^3$ & non-rigid component (object motion) of $\mathbf{F}^i$ \\
        $f \in \mathbb{R}^+$ & focal length (assumed identical on both axes) \\
        $\mathbf{c} \in \mathbb{R}^2$ & optical center of the camera \\
        $\mathbf{p} \in \mathbb{R}^{H\times W \times 2}$ & projected point position map for $\mathbf{I}$ \\
        $\mathbf{f} \in \mathbb{R}^{H\times W \times 2}$ & optical flow map for $\mathbf{I}$ \\
    \bottomrule
    \end{tabular}
\end{table}

For reproducibility, we also list the key optimization and schedule settings used in the two-stage training pipeline. Table~\ref{tab:train_hparams} reports the main hyperparameters shared across experiments.

\begin{table}[h]
    \caption{Key training hyperparameters.}
    \label{tab:train_hparams}
    \centering
    \setlength{\tabcolsep}{5pt}
    \begin{tabular}{lcc}
    \toprule
        \textbf{Setting} & \textbf{Stage 1} & \textbf{Stage 2} \\
        \midrule
        Epochs & 100 & 100 \\
        Resolution & 224 & 512 \\
        Pairs per epoch & 900K & 84K \\
        Batch size & 256 & 64 \\
        Warmup epochs & 10 & 20 \\
        Peak learning rate & \multicolumn{2}{c}{1e-4} \\
        Final learning rate & \multicolumn{2}{c}{1e-6} \\
        Optimizer & \multicolumn{2}{c}{Adam} \\
        LR schedule & \multicolumn{2}{c}{linear warmup + cosine decay} \\
        Gradient clipping & \multicolumn{2}{c}{max norm 10} \\
    \bottomrule
    \end{tabular}
\end{table}

\end{document}